\documentclass[letterpaper, 10pt, conference]{ieeeconf}

\IEEEoverridecommandlockouts
\usepackage{amsmath}
\usepackage{amssymb}
\usepackage{amsfonts}
\usepackage{mathtools}

\usepackage{graphicx}

\usepackage{booktabs}
\usepackage{multirow}

\usepackage{bbm}
\usepackage{dblfloatfix}
\usepackage{placeins}
\usepackage{xcolor}
\usepackage{hyperref}

\title{\LARGE \bf
TRIAGE: \underline{T}ype-\underline{R}outed \underline{I}nterventions via \underline{A}leatoric-Epistemic\\
\underline{G}ated \underline{E}stimation in Robotic Manipulation and Adaptive Perception\\[6pt]
\large Don't Treat All Uncertainty the Same}

\author{Divake Kumar$^{*1}$, Sina Tayebati$^{1}$, Devashri Naik$^{1}$, Patrick Poggi$^{1}$, Amanda Sofie Rios$^{2}$\\
Nilesh Ahuja$^{2}$, Amit Ranjan Trivedi$^{1}$\\
$^{1}$University of Illinois at Chicago \quad $^{2}$Intel Labs}

\makeatletter
\let\old@maketitle\@maketitle
\def\@maketitle{%
  \old@maketitle
  \vskip 0.5em
  \begin{center}
  \includegraphics[width=\textwidth]{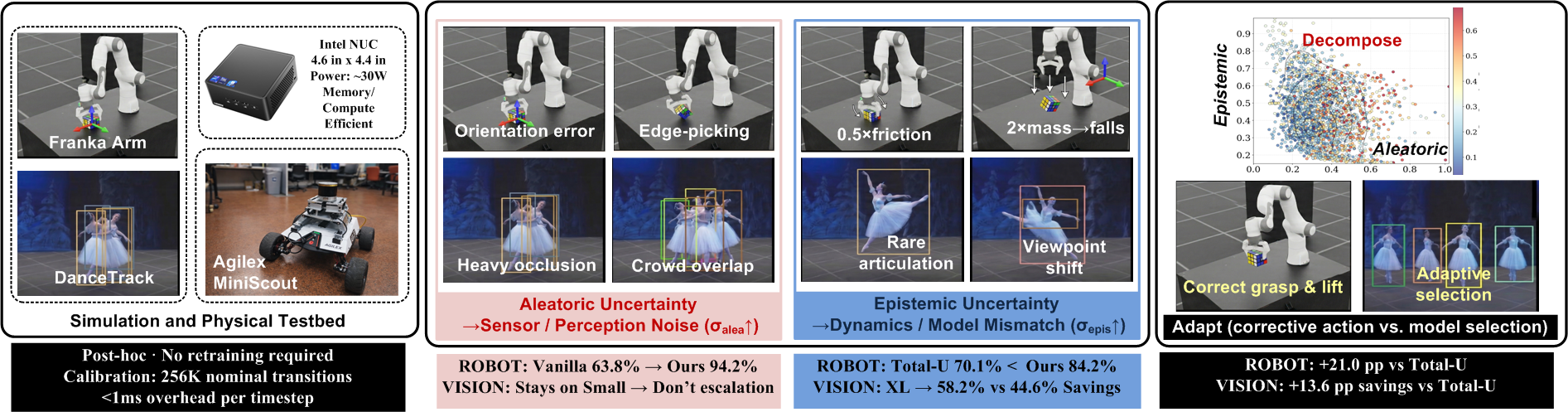}
  \end{center}
  \vskip 2pt
  {\small\noindent\textbf{Fig.~1.}\ Overview of the proposed uncertainty decomposition across robotics and visual perception.
  \textbf{Top:} Aleatoric uncertainty reflects sensor or perception noise (e.g., orientation error, occlusion), while epistemic uncertainty reflects dynamics or representation mismatch (e.g., friction change, viewpoint shift). Examples are shown for Franka manipulation and DanceTrack tracking.
  \textbf{Bottom:} Observations processed on an edge platform produce aleatoric and epistemic signals that remain nearly orthogonal ($r=0.048$, $21{,}324$ detections). These signals guide corrective responses in control and adaptive model selection in tracking. The approach improves manipulation robustness (94.2\% vs.\ 63.8\%) and reduces tracking compute by 58.2\% on MOT17 with negligible accuracy loss.\par}
  \refstepcounter{figure}\label{fig:overview}
  
}
\makeatother

\begin{document}

\maketitle
\thispagestyle{empty}
\pagestyle{empty}

\begin{abstract}
Most uncertainty-aware robotic systems collapse prediction uncertainty into a single scalar score and use it to trigger uniform corrective responses. This aggregation obscures whether uncertainty arises from corrupted observations or from mismatch between the learned model and the true system dynamics. As a result, corrective actions may be applied to the wrong component of the closed loop, degrading performance relative to leaving the policy unchanged. We introduce a lightweight \textit{post hoc} framework that decomposes uncertainty into aleatoric and epistemic components and uses these signals to regulate system responses at inference time. Aleatoric uncertainty is estimated from deviations in the observation distribution using a Mahalanobis density model, while epistemic uncertainty is detected using a noise robust forward dynamics ensemble that isolates model mismatch from measurement corruption. The two signals remain empirically near orthogonal during closed loop execution and enable type specific responses. High aleatoric uncertainty triggers observation recovery, while high epistemic uncertainty moderates control actions. The same signals also regulate adaptive perception by guiding model capacity selection during tracking inference. Experiments demonstrate consistent improvements across both control and perception tasks. In robotic manipulation, the decomposed controller improves task success from 59.4\% to 80.4\% under compound perturbations and outperforms a combined uncertainty baseline by up to 21.0\%. In adaptive tracking inference on MOT17, uncertainty-guided model selection reduces average compute by 58.2\% relative to a fixed high capacity detector while preserving detection quality within 0.4\%. Code and demo videos are available at \href{https://divake.github.io/uncertainty-decomposition/}{\textcolor{blue}{\texttt{divake.github.io/uncertainty-decomposition}}}.
\end{abstract}

\section{Introduction}
\label{sec:intro}

Prediction uncertainty remains poorly handled in closed loop robot control systems. Most uncertainty-aware controllers collapse prediction error into a single scalar quantity and use it to trigger a uniform conservative response \cite{chua2018deep,lutjens2019safe,wu2022uncertainty}. This aggregation obscures whether uncertainty arises from corrupted measurements or from mismatch between the learned dynamics model and the true environment. As a result, the controller cannot determine \textit{why} it is uncertain, \textit{what} corrective actions are principally aligned with the underlying disturbance, and \textit{how} to resolve this ambiguity.

In many robotic systems two distinct perturbation mechanisms arise. \emph{Sensor perturbation} corrupts observations through encoder noise, calibration drift, or interference while the underlying plant dynamics remain unchanged. \emph{Dynamics shift} alters the physical response of the system, for example when object mass changes, friction varies, or actuator characteristics drift, thereby modifying the relationship between control input and state transition. These mechanisms occur across manipulation, locomotion, and trajectory tracking tasks where training and deployment conditions may differ.

Because these perturbations affect different components of the control loop, they require different corrective responses: sensor perturbations require improving the observation without modifying the control input, while dynamics shifts require adjusting the control action while leaving the observation unchanged. This distinction aligns naturally with the separation between \emph{aleatoric} and \emph{epistemic} uncertainty \cite{kendall2017uncertainties,hullermeier2021aleatoric,der2009aleatory}. The importance of principled uncertainty handling extends beyond robotics to language and vision-language models, where conformal abstention policies have been shown to improve risk management under distribution shift~\cite{tayebati2025learning}. Aleatoric uncertainty reflects stochastic variation in observations that affects state estimation, whereas epistemic uncertainty reflects model mismatch or distributional shift that affects action consistency with the true system dynamics. Collapsing these quantities into a single score removes structural information about the origin of the disturbance and therefore about the appropriate corrective action.

Addressing this, we develop a \textit{post hoc} uncertainty decomposition framework for robot control that preserves this structure. Aleatoric uncertainty is estimated using a Mahalanobis distance in observation space \cite{mahalanobis1936generalized,lee2018simple}, measuring deviation from the nominal state distribution. Epistemic uncertainty is estimated using a noise-robust dynamics ensemble trained on clean and noise-augmented transitions with clean targets, allowing dynamics mismatch to be distinguished from sensor corruption. Across nominal and perturbed rollouts the resulting signals exhibit low empirical correlation, enabling intervention rules aligned with the disturbance source. High aleatoric uncertainty triggers observation recovery, while high epistemic uncertainty triggers action dampening, thus preserving consistency between estimation correction and control adaptation without perturbing the original policy.

We evaluate the framework across robotic manipulation and trajectory tracking tasks. Under compound perturbations the decomposed controller achieves 80.4\% task success compared to 59.4\% for a combined-uncertainty baseline and 44.6\% for the vanilla policy. The combined baseline performs below the vanilla controller under every tested dynamics shift condition, indicating that mismatched corrective actions can degrade performance. The decomposed controller remains stable across a $3.3\times$ range of dampening gains, whereas the combined baseline degrades monotonically. In trajectory tracking, the same decomposition enables adaptive model selection for visual detection, reducing compute by 58.2\% on MOT17 while maintaining detection quality within 0.4\% of the largest backbone. These results show that separating uncertainty sources improves both corrective control and inference-time resource allocation. Our contributions are:
\begin{itemize}

\item \textbf{Principled uncertainty decomposition for closed-loop decision making.}
A \textit{post hoc} framework that separates aleatoric and epistemic uncertainty and maps them to distinct system responses, enabling corrective actions that remain consistent with the underlying disturbance.

\item \textbf{Robust detection of dynamics mismatch.}
A noise-robust dynamics ensemble that distinguishes dynamics shift from sensor corruption by training with noise-augmented inputs.

\item \textbf{Unified use across control and perception.}
We show that the same uncertainty decomposition supports both corrective control in robotic manipulation and adaptive model selection for visual tracking.

\item \textbf{Structural limitations of monolithic uncertainty.}
We show that collapsing uncertainty into a single scalar can reduce performance below the vanilla controller under dynamics shift, while the proposed decomposition improves manipulation robustness and reduces tracking compute by 58.2\% on MOT17 with negligible loss.

\end{itemize}

\section{Related Work}
\label{sec:related}

\noindent\textbf{Aleatoric and epistemic uncertainty.}
The distinction between aleatoric uncertainty arising from data variability and epistemic uncertainty arising from model uncertainty is central in probabilistic machine learning~\cite{der2009aleatory,hullermeier2021aleatoric}. Kendall and Gal~\cite{kendall2017uncertainties} formalized this decomposition for deep networks, modeling aleatoric uncertainty through learned heteroscedastic variance and epistemic uncertainty via MC Dropout~\cite{gal2016dropout}. Lakshminarayanan et al.~\cite{lakshminarayanan2017simple} showed that deep ensembles provide strong uncertainty estimates without explicit Bayesian inference. Subsequent analyses have examined limitations of this decomposition~\cite{valdenegro2022deeper}, yet the practical consensus is that the two sources reflect distinct failure mechanisms. Recent work has combined conformal prediction with evidential learning to provide distribution-free guarantees while preserving aleatoric--epistemic separability~\cite{stutts2024conformal}, and calibrated decomposition of uncertainty in deep features has been shown to enable inference-time adaptation~\cite{kumar2025calibrated}. We adopt this distinction for closed-loop robot control, where conflating uncertainty types leads to physically misaligned interventions.

\vspace{3pt}
\noindent\textbf{Uncertainty in model-based control.}
Ensemble dynamics models are widely used for uncertainty-aware planning. PETS~\cite{chua2018deep} propagates trajectories through probabilistic ensembles to capture predictive uncertainty during model predictive control, but aggregates uncertainty into a single signal for conservative planning. Nagabandi et al.~\cite{nagabandi2018neural} employ neural dynamics models for model-based reinforcement learning with model-free refinement, without decomposing prediction error sources. Das et al.~\cite{das2023uncertainty} propose real-time uncertainty decomposition for nonlinear control using an OOD classifier for epistemic uncertainty and a learned variance head for aleatoric uncertainty; however, their epistemic estimator relies on labeled OOD data during training. L{\"u}tjens et al.~\cite{lutjens2019safe} use ensemble disagreement for safe reinforcement learning but do not translate uncertainty into type-specific corrective actions. In these approaches, uncertainty typically modulates conservatism uniformly by slowing actions or penalizing reward, without distinguishing whether it arises from corrupted observations or dynamics mismatch. In contrast, our framework uses noise-augmented training to actively decouple sensor noise from dynamics mismatch in the epistemic estimator, maps each uncertainty type to a distinct corrective action rather than modulating a single conservatism parameter, and operates \textit{post hoc} on a frozen policy without requiring retraining or labeled OOD data. Recent work on uncertainty-aware edge robotics~\cite{darabi2024navigating,trivedi2025intelligent} and conformal prediction for autonomous perception~\cite{kumar2025lidar,kumar2025learnable} has highlighted the need for structured uncertainty signals in resource-constrained settings; our approach provides such structure through lightweight post hoc decomposition.

\vspace{3pt}
\noindent\textbf{Uncertainty for Multi Object Tracking:} Tracking by detection methods such as DeepSORT \cite{wojke2017simple}, FairMOT \cite{zhang2021fairmot}, ByteTrack \cite{zhang2022bytetrack}, and transformer approaches \cite{zhang2023motrv2,gao2023memotr,cai2022memot} base reliability primarily on detector confidence. This merges observation noise and representation mismatch into one scalar and limits adaptation to changing video conditions. Existing uncertainty aware tracking approaches use entropy or regression variance, which remain single mode indicators. Current methods do not connect structured uncertainty to inference time decisions, though dynamic networks \cite{han2021dynamic,teerapittayanon2016branchynet,bolukbasi2017adaptive} enable adaptive computation. Comparatively, our formulation separates normalized components that distinguish data quality from representation support, enabling context expansion or adaptive model switching. Concurrent work on uncertainty-guided depth adaptation for transformer-based tracking~\cite{poggi2026uncertainty} demonstrates similar benefits of uncertainty-driven inference-time decisions in visual tracking.

\vspace{3pt}
\noindent\textbf{Robustness through domain randomization.}
Domain randomization~\cite{tobin2017domain} and dynamics randomization~\cite{peng2018sim} train policies across environment distributions so that deployment conditions appear as in-distribution variations. This strategy has enabled sim-to-real transfer in dexterous manipulation~\cite{andrychowicz2020learning} and legged locomotion~\cite{lee2020learning,hwangbo2019learning}. Rapid Motor Adaptation~\cite{kumar2021rma} conditions policies on recent history to infer latent dynamics online. These methods build robustness during training and require exposure to distributional variation beforehand. Noise-tolerant training through adversarial curriculum methods~\cite{darabi2025intact} also improves perception robustness but requires retraining the model. Our approach is complementary: it operates \textit{post hoc} on a fixed policy with only a brief nominal calibration rollout, extending reliability to perturbations outside the training distribution.

\begin{figure*}[t]
\centering
\begin{tabular}{ccc}
\includegraphics[height=4.2cm]{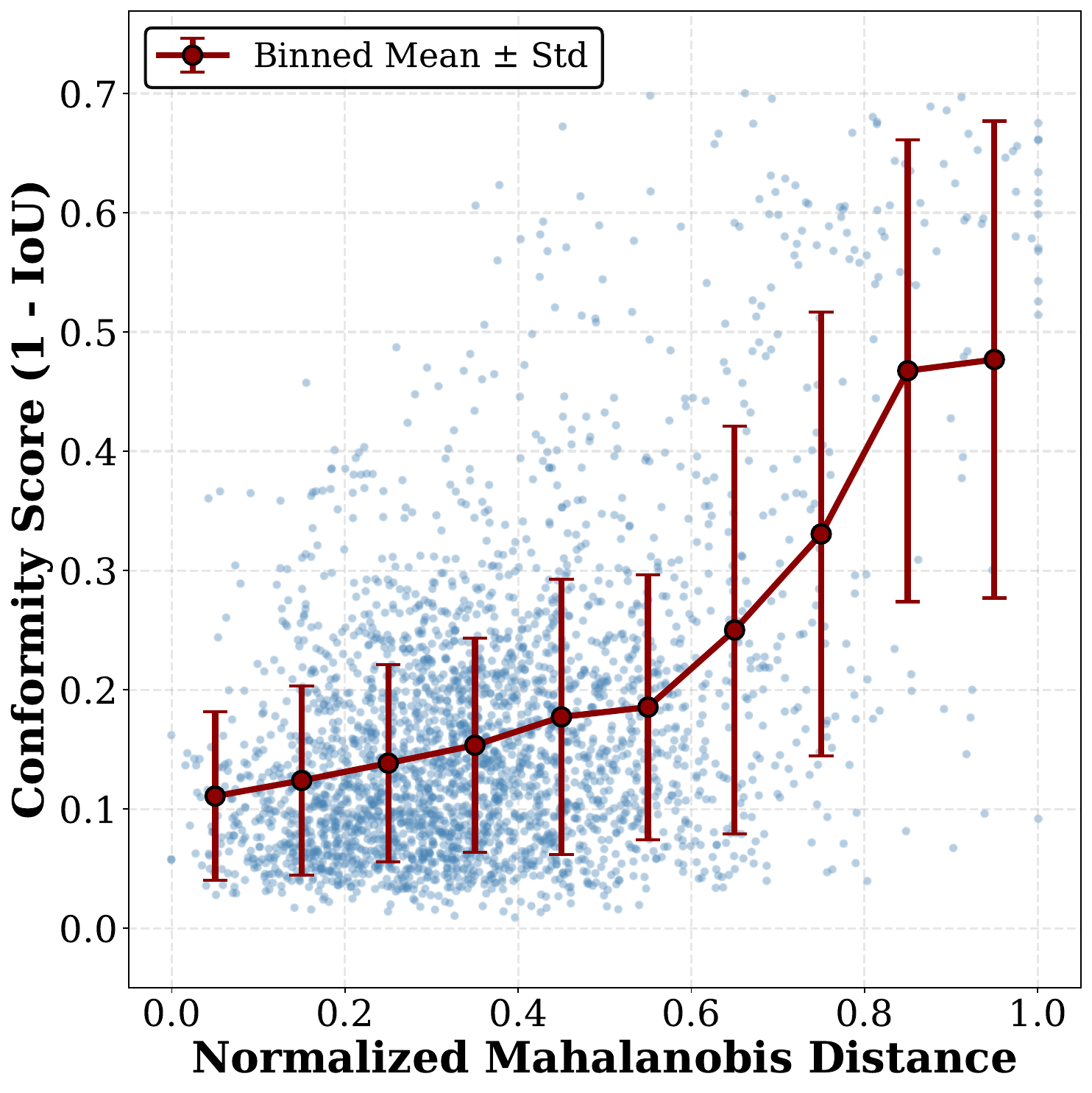} &
\includegraphics[height=4.2cm]{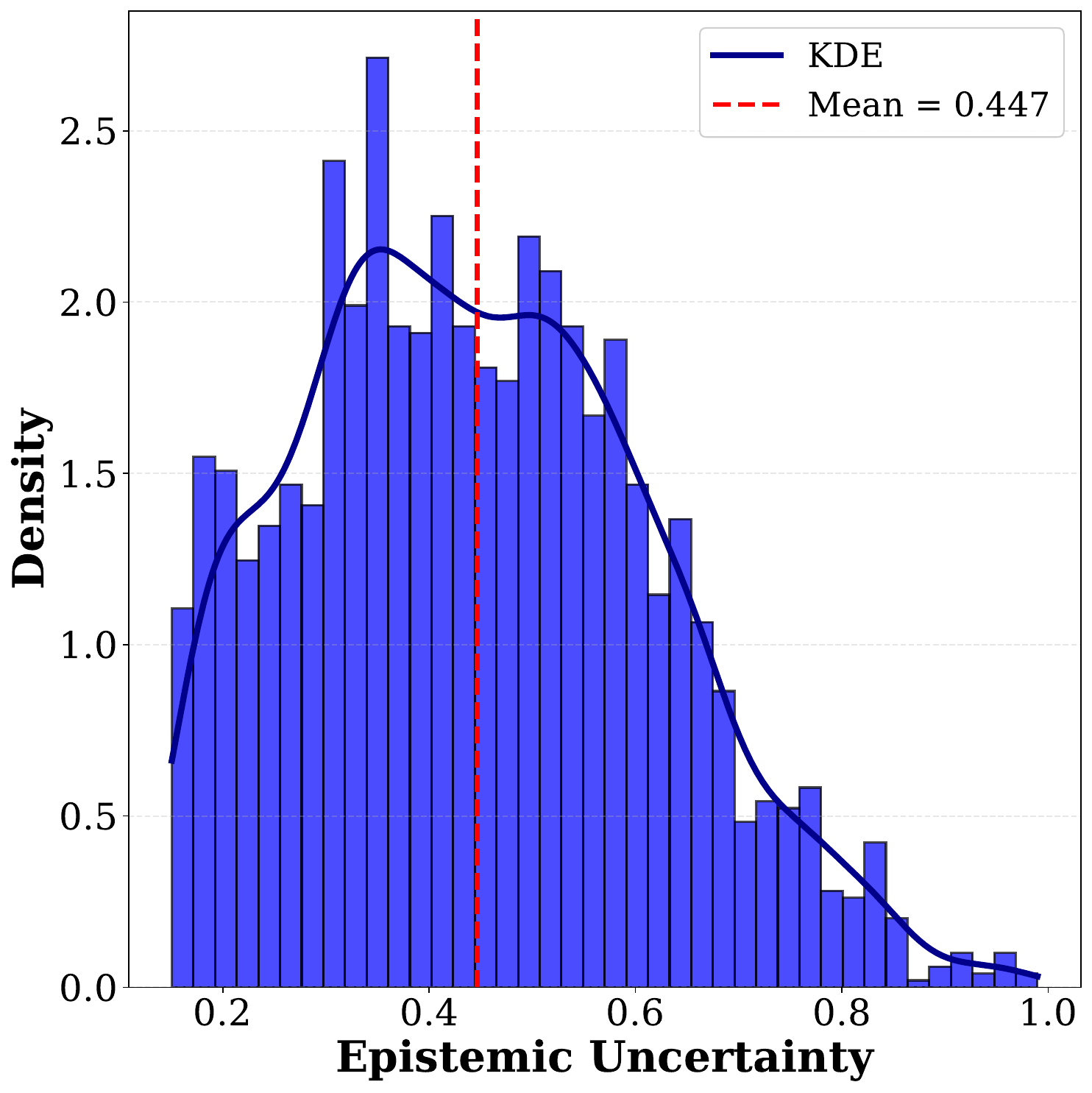} &
\includegraphics[height=4.2cm]{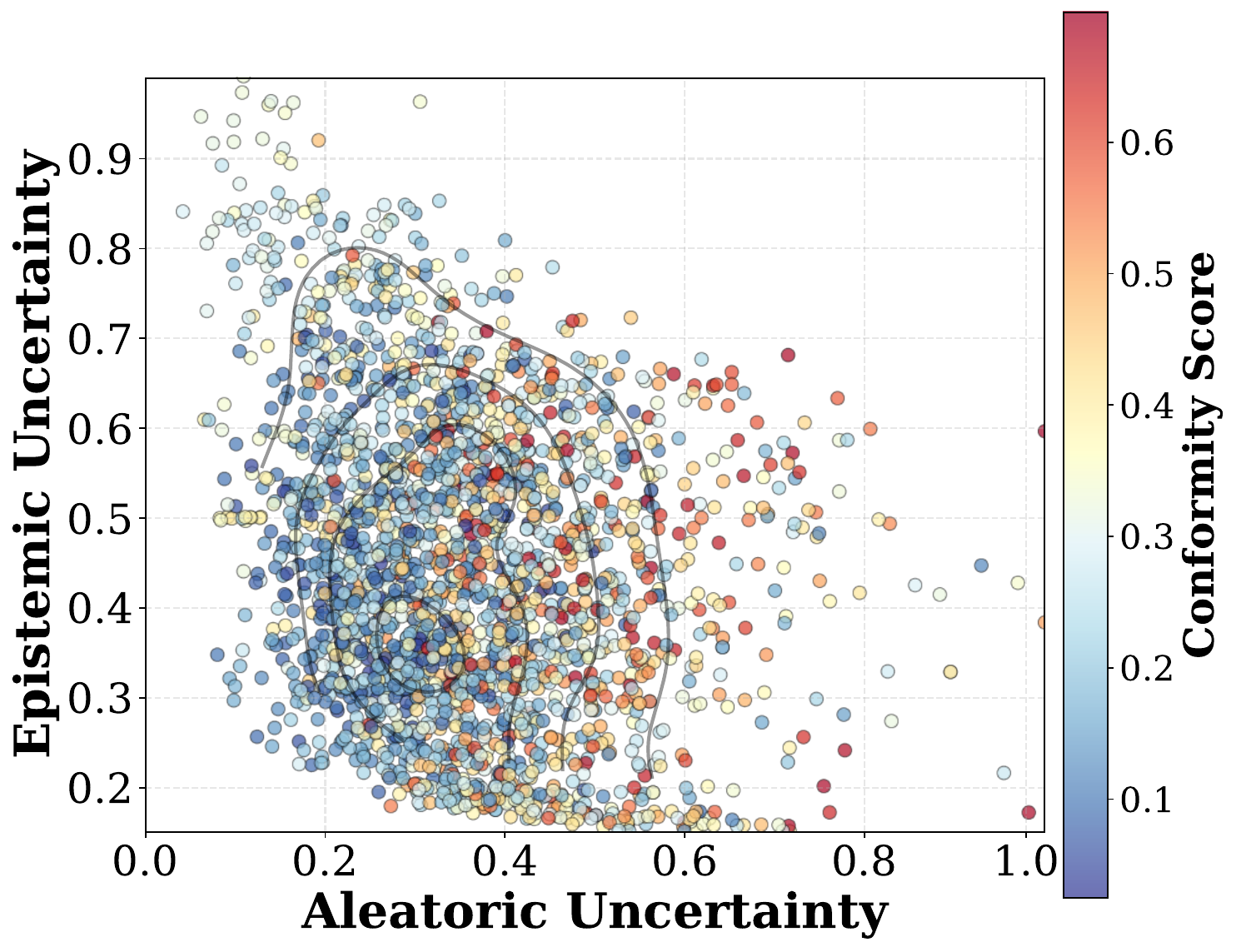} \\
(a) Aleatoric validation & (b) Epistemic distribution & (c) Orthogonality
\end{tabular}

\caption{
Empirical behavior of uncertainty signals on MOT17.
(a) Aleatoric uncertainty correlates with prediction error, indicating sensitivity to observation corruption.
(b) Epistemic uncertainty exhibits a broader spread reflecting variation in representation support.
(c) Joint distribution of aleatoric and epistemic signals across 21,324 detections shows near-zero correlation ($r=0.048$), confirming that the two uncertainties capture distinct disturbance mechanisms.
}

\label{fig:uncert_behavior}
\end{figure*}

\section{Decomposed Uncertainty for Closed-Loop Robot Control}
\label{sec:framework}

A robotic controller $\pi$ maps observations $\mathbf{o}_t \in \mathbb{R}^d$ to control inputs $\mathbf{a}_t$. As described in Sec.~\ref{sec:intro}, sensor perturbations and dynamics shifts affect different components of the closed loop and require different corrective responses. We estimate two uncertainty signals that preserve this structure: aleatoric uncertainty captures deviations in the observation distribution and gates observation-level recovery, while epistemic uncertainty captures dynamics mismatch and regulates control adaptation.

This decomposition is grounded in the structure of the control loop itself. Sensor perturbations affect the observation $\mathbf{o}_t$ but leave the true transition kernel $p(\mathbf{s}_{t+1} \mid \mathbf{s}_t, \mathbf{a}_t)$ unchanged, whereas dynamics shifts alter the transition kernel while leaving the sensor model intact. The Mahalanobis distance measures deviation from the nominal observation distribution and is therefore sensitive to the first mechanism by construction. Under the Gaussian model it is a sufficient statistic for the observation log-density~\cite{lee2018simple}, making it the natural measure of observation quality. The dynamics ensemble prediction error measures discrepancy between the learned and true transition functions and is therefore sensitive to the second mechanism by construction. Noise-augmented training (Eq.~\eqref{eq:ensemble_train}) further enforces this separation by making the ensemble invariant to observation-level corruption, so that prediction error increases only when the underlying dynamics change. This yields an operational definition of epistemic uncertainty that does not require a posterior over model parameters: a fixed ensemble serves as a reference model, and prediction error at deployment time provides the epistemic signal without parameter updates or sampling.

\vspace{4pt}
\noindent\textbf{Calibration via nominal rollouts.}
Both estimators require a reference distribution representing nominal system behavior. We execute the trained controller under nominal conditions for $N_{\text{cal}}$ timesteps and collect transition tuples
\[
\mathcal{D}_{\text{cal}} =
\bigl\{(\mathbf{o}_t^{(i)},\, \mathbf{a}_t^{(i)},\, \mathbf{o}_{t+1}^{(i)})\bigr\}_{i=1}^{N_{\text{cal}}},
\]
which characterize the observation distribution encountered during successful operation together with the corresponding state transitions.

\vspace{4pt}
\noindent\textbf{Aleatoric Estimation via Observation Density.}
Sensor perturbations displace observations from their nominal distribution. 
To detect such deviations we estimate the empirical mean and regularized covariance of calibration observations over active dimensions $\mathcal{A}$, excluding constant components such as goal parameters:
\[
\boldsymbol{\mu} =
\frac{1}{N_{\text{cal}}}
\sum_{i=1}^{N_{\text{cal}}} \mathbf{o}_{\mathcal{A}}^{(i)},
\qquad
\boldsymbol{\Sigma} =
\frac{1}{N_{\text{cal}}}
\sum_{i=1}^{N_{\text{cal}}}
(\mathbf{o}_{\mathcal{A}}^{(i)} - \boldsymbol{\mu})
(\mathbf{o}_{\mathcal{A}}^{(i)} - \boldsymbol{\mu})^\top
+ \lambda \mathbf{I},
\]
where $\lambda = 10^{-6}$ ensures numerical stability.

At runtime the aleatoric score for a new observation $\mathbf{o}_t$ is computed as the Mahalanobis distance
\begin{equation}
\sigma_{\text{alea}}(\mathbf{o}_t)
=
\sqrt{
(\mathbf{o}_{t,\mathcal{A}} - \boldsymbol{\mu})^\top
\boldsymbol{\Sigma}^{-1}
(\mathbf{o}_{t,\mathcal{A}} - \boldsymbol{\mu})
}.
\label{eq:alea}
\end{equation}

The detection threshold $\tau_{\text{alea}}$ is defined as the 95th percentile of calibration scores, optionally scaled by a multiplier $m_a$ (we use $m_a = 1.0$). 
Observations that fall outside the nominal therefore produce elevated $\sigma_{\text{alea}}$ values, indicating likely sensor corruption.
Fig.~\ref{fig:uncert_behavior}(a) shows the empirical relationship between the aleatoric score and prediction error on the calibration set. 
Higher values correspond to larger residual error, confirming that the estimator captures observation corruption rather than dynamics mismatch.

\vspace{4pt}
\noindent\textbf{Epistemic Estimation via Dynamics Prediction.}
Dynamics shifts alter the system response while sensor measurements remain accurate. 
To detect such shifts, we learn a forward dynamics model from nominal calibration data and monitor prediction error during deployment. When the true system dynamics deviate from the learned model, predicted and observed state transitions diverge. We train an ensemble of $K{=}5$ forward dynamics MLPs that predict the observation change between consecutive timesteps:
\begin{equation}
f_k(\mathbf{o}_{t,\mathcal{A}},\, \mathbf{a}_t)
\approx
\Delta\mathbf{o}_{t+1,\mathcal{A}}
=
\mathbf{o}_{t+1,\mathcal{A}} - \mathbf{o}_{t,\mathcal{A}},
\quad k=1,\ldots,K.
\label{eq:dynamics_model}
\end{equation}

Training only on clean observations would cause prediction error to increase under sensor noise, confounding measurement perturbations with dynamics shifts. 
To mitigate this effect, we apply noise-augmented training. Each model is trained on clean, moderately noisy ($1\times$), and heavily noisy ($2\times$) observations, all paired with the same clean transition target $\Delta\mathbf{o}_{t+1}$:
\begin{equation}
\min_{\theta_k}
\sum_{(\tilde{\mathbf{o}}_t,\, \mathbf{a}_t,\, \Delta\mathbf{o}_{t+1}) \in \tilde{\mathcal{D}}}
\left\|
f_k(\tilde{\mathbf{o}}_{t,\mathcal{A}}, \mathbf{a}_t;\theta_k)
-
\Delta\mathbf{o}_{t+1,\mathcal{A}}
\right\|^2.
\label{eq:ensemble_train}
\end{equation}

This training procedure encourages the ensemble to recover the underlying system dynamics despite observation noise. 
Consequently, sensor perturbations alone produce only limited prediction error increases, whereas genuine dynamics shifts lead to larger prediction discrepancies.

At timestep $t$, the epistemic signal is computed by comparing the ensemble mean prediction with the observed state transition:
\begin{equation}
\sigma_{\text{epis}}(\mathbf{o}_{t-1}, \mathbf{a}_{t-1}, \mathbf{o}_t)
=
\sqrt{
\frac{1}{|\mathcal{G}|}
\sum_{j \in \mathcal{G}}
\Bigl(
\bar{f}_j(\mathbf{o}_{t-1,\mathcal{A}}, \mathbf{a}_{t-1})
-
\Delta o_{t,j}
\Bigr)^2
},
\label{eq:epis}
\end{equation}
where $\bar{f}_j = \frac{1}{K}\sum_{k} f_{k,j}$ denotes the ensemble mean prediction for dimension $j$, and $\mathcal{G}$ is the subset of reliably predicted dimensions identified during calibration ($R^2 > 0.3$). Restricting evaluation to $\mathcal{G}$ prevents inherently noisy or poorly modeled dimensions from inflating epistemic signal.

Prediction errors observed during closed-loop execution are typically larger than those measured offline because trajectories induce strong temporal correlations. Static thresholds therefore lead to frequent false positives. To compensate for this effect, we perform a short nominal rollout of $T_{\text{cal}}{=}300$ steps and set the epistemic threshold as
\begin{equation}
\tau_{\text{epis}}
=
Q_{95}\bigl(\{\sigma_{\text{epis}}^{(t)}\}_{t=1}^{T_{\text{cal}}}\bigr)
\times m_e,
\label{eq:runtime_cal}
\end{equation}
where $Q_{95}$ denotes the 95th percentile and $m_e{=}1.5$ is a scaling factor. 
This runtime calibration maintains low trigger rates under nominal operation while preserving sensitivity to dynamics shifts. Fig.~\ref{fig:uncert_behavior}(b) shows the distribution of epistemic uncertainty across detections. 
The broader spread reflects variation in representation support and indicates sensitivity to dynamics mismatch rather than observation corruption. Fig.~\ref{fig:uncert_behavior}(c) shows the joint distribution of $\sigma_{\text{alea}}$ and $\sigma_{\text{epis}}$, revealing near-zero correlation ($r{=}0.048$) between the two signals. This empirical near-orthogonality is consistent with the structural argument above: observation density and dynamics prediction error measure distinct physical quantities in the control loop, so they respond independently to different disturbance sources.

\vspace{4pt}
\noindent\textbf{Type-Specific Interventions.}
The uncertainty decomposition produces two signals that correspond to distinct disturbance mechanisms and therefore trigger different responses.

\vspace{3pt}
\noindent\textit{Observation recovery when $\sigma_{\text{alea}} > \tau_{\text{alea}}$.}
When the aleatoric estimator indicates sensor degradation, we recover a cleaner observation using the simulator observation pipeline. In Isaac Lab~\cite{mittal2023orbit}, observations are generated in two stages: the physics engine first produces a noiseless system state, after which a sensor model injects measurement noise. When $\sigma_{\text{alea}}$ exceeds its threshold, we resample the sensor model $N$ times from the same physics state and average the resulting observations:
\begin{equation}
\hat{\mathbf{o}}_t
=
\frac{1}{N}
\sum_{n=1}^{N}
\bigl(
\mathbf{o}_t^{\text{phys}} + \boldsymbol{\epsilon}_n
\bigr),
\qquad
\boldsymbol{\epsilon}_n \sim \mathcal{N}(\mathbf{0}, \boldsymbol{\Sigma}_{\text{sensor}}).
\label{eq:denoise}
\end{equation}

For zero-mean Gaussian noise the observation variance decreases by a factor of $N$; we use $N{=}5$. This oracle recovery mechanism isolates the effect of the uncertainty decomposition from the choice of denoising method; in physical systems the same trigger could activate Kalman filtering or a learned denoiser.

\vspace{3pt}
\noindent\textit{Action dampening when $\sigma_{\text{epis}} > \tau_{\text{epis}}$.}
When the epistemic estimator detects dynamics shift, the learned policy may produce actions that are inconsistent with the true system response. To mitigate this effect we reduce the control magnitude
\begin{equation}
\mathbf{a}_t' = (1 - \alpha)\mathbf{a}_t,
\qquad
\alpha \in [0,1],
\label{eq:dampen}
\end{equation}
with $\alpha = 0.30$ unless stated otherwise. Dampening reduces the impact of model mismatch while maintaining the policy structure, allowing the controller to remain stable under altered dynamics.

\vspace{4pt}
\noindent\textbf{Decomposed Control Policy.}
The two interventions operate on different components of the control loop and are triggered independently according to the estimated uncertainty signals:
\begin{align}
\sigma_{\text{alea}}(\mathbf{o}_t) > \tau_{\text{alea}}
&\;\Longrightarrow\;
\text{apply observation recovery}, \label{eq:dec_alea} \\
\sigma_{\text{epis}}(\mathbf{o}_{t-1}, \mathbf{a}_{t-1}, \mathbf{o}_t) > \tau_{\text{epis}}
&\;\Longrightarrow\;
\text{apply action dampening}. \label{eq:dec_epis}
\end{align}

Under sensor perturbation only Eq.~\eqref{eq:dec_alea} activates, whereas under dynamics shift only Eq.~\eqref{eq:dec_epis} activates. Under compound perturbations both interventions may activate concurrently, each addressing its corresponding disturbance source.

For comparison, the \emph{Total-U} baseline aggregates the two uncertainty signals using
\[
(\sigma_{\text{alea}} > \tau_{\text{alea}})
\lor
(\sigma_{\text{epis}} > \tau_{\text{epis}}),
\]
and applies both interventions whenever either signal fires. Under pure dynamics shift this triggers unnecessary observation recovery alongside dampening, producing mismatched responses.

\vspace{4pt}
\noindent\textbf{Experimental validation.}
We evaluate this principle in two complementary settings. Section~\ref{sec:robo_experiments} studies corrective control in robotic manipulation, where the decomposition regulates observation recovery and action adaptation under sensor perturbations and dynamics shifts. Section~\ref{sec:tracking} examines inference-time decision making in visual tracking, where the same uncertainty signals guide adaptive model selection. 
\section{Robotic Manipulation Experiments}
\label{sec:robo_experiments}

\subsection{Experimental Setup}

\noindent\textbf{Task.}
We evaluate on \texttt{Isaac-Lift-Cube-Franka-v0} in NVIDIA Isaac Lab~\cite{mittal2023orbit}. A 7-DOF Franka Emika Panda arm must reach, grasp, and lift a cube to a target height. Observations are 36-dimensional and include joint positions (9), joint velocities (9), object position (3), target pose (7), and previous actions (8). The controller is a pre-trained MLP policy with running-mean observation normalization.

\vspace{3pt}
\noindent\textbf{Perturbation conditions.}
We evaluate four conditions that isolate and combine disturbance types:
\emph{Nominal}, with no perturbation;
\emph{Sensor perturbation}, where Gaussian noise is applied to joint positions ($\sigma{=}0.05$), joint velocities ($\sigma{=}0.1$), and object position ($\sigma{=}0.05$);
\emph{Dynamics shift}, where object mass or surface friction is modified;
and \emph{Compound perturbation}, where both perturbations are applied simultaneously.

\vspace{3pt}
\noindent\textbf{Evaluation protocol.}
Experiments run 1{,}000 parallel simulation environments. Each method is evaluated on 1{,}000 independent episodes per perturbation condition. Calibration uses 256K nominal transitions collected under the unperturbed policy.

\vspace{3pt}
\noindent\textbf{Methods.}
We compare five controllers:
(i) \emph{Vanilla}, the unmodified policy;
(ii) \emph{Observation Recovery Only}, which applies observation recovery at every timestep;
(iii) \emph{Dampen Only}, which applies action dampening at every timestep;
(iv) \emph{Total-U}, which aggregates uncertainty signals and applies both interventions whenever either trigger activates;
(v) \emph{Decomposed} (ours), which activates observation recovery and action dampening independently according to their respective uncertainty signals.

\begin{table}[t]
\caption{Task success rate (\%) on Isaac-Lift-Cube-Franka under different perturbation conditions ($N{=}1000$ episodes per method). Dynamics shift uses $2\times$ object mass. $\Delta$ reports improvement over Total-U.}
\label{tab:robo_main}
\centering
\small
\setlength{\tabcolsep}{2pt}
\begin{tabular}{lcccc}
\toprule
\textbf{Method} & \textbf{Nominal} & \textbf{Sensor} & \textbf{Dynamics} & \textbf{Compound} \\
\midrule
Vanilla                    & 100.0 & 63.8 & 72.4 & 44.6 \\
Observation Recovery  & 97.4  & 89.2 & 64.8 & 57.3 \\
Dampen Only                & 88.6  & 58.4 & 78.3 & 50.2 \\
Total-U (combined)         & 96.8  & 80.6 & 70.1 & 59.4 \\
\midrule
\textbf{Decomposed (ours)} & \textbf{99.6} & \textbf{94.2} & \textbf{84.2} & \textbf{80.4} \\
\midrule
$\Delta$ vs.\ Total-U      & +2.8 & +13.6 & +14.1 & +21.0 \\
\bottomrule
\end{tabular}
\end{table}

\subsection{Results}

Table~\ref{tab:robo_main} reports task success with a $2\times$ object mass perturbation representing dynamics shift. The $\Delta$ row summarizes the main result. The performance gap between \emph{Decomposed} and \emph{Total-U} increases with perturbation complexity: $+2.8$, $+13.6$, $+14.1$, and $+21.0$~pp under nominal, sensor, dynamics, and compound conditions respectively. The difference is small under nominal conditions, indicating negligible overhead. Under single perturbations the advantage increases to 13--14~pp, and under compound perturbations it reaches 21~pp. These results show that separating disturbance types produces more effective responses than a single aggregated uncertainty signal. 

\vspace{3pt}
\noindent\textbf{Intervention overhead under nominal conditions.}
The nominal column measures the cost of applying interventions when no perturbation is present. \emph{Dampen Only} achieves 88.6\% because reducing control magnitude weakens feedback during borderline grasps. \emph{Observation Recovery Only} reaches 97.4\% due to resampling artifacts applied to already accurate observations. \emph{Total-U} achieves 96.8\% because combined triggers occasionally activate both interventions near threshold boundaries. In contrast, \emph{Decomposed} maintains 99.6\% success since both uncertainty signals remain below threshold.

\vspace{3pt}
\noindent\textbf{Sensor perturbation.}
Under sensor perturbation \emph{Decomposed} achieves 94.2\% success by applying observation recovery only. \emph{Dampen Only} drops to 58.4\%, below \emph{Vanilla} at 63.8\%. Reducing control magnitude does not correct corrupted observations and therefore degrades performance. \emph{Total-U} reaches 80.6\%, higher than \emph{Dampen Only} but lower than \emph{Observation Recovery Only} at 89.2\% because unnecessary dampening reduces performance. In the decomposed controller the epistemic signal rarely activates under sensor perturbation, with fewer than 2\% of timesteps triggering dampening. The controller therefore applies observation recovery while preserving the original control input.

\vspace{3pt}
\noindent\textbf{Dynamics shift.}
Under dynamics shift \emph{Total-U} achieves 70.1\%, below \emph{Vanilla} at 72.4\%. The combined trigger applies observation recovery to already correct sensor readings on 17.2\% of timesteps, introducing resampling artifacts that degrade the state estimate and offset the benefit of dampening. \emph{Observation Recovery Only} shows the same effect with 64.8\%, 7.6~pp below \emph{Vanilla}. These results show that observation correction under dynamics shift can reduce performance. \emph{Decomposed} avoids this failure mode. Under dynamics shift the epistemic signal activates while the aleatoric signal remains near its nominal false positive rate, yielding 84.2\% success.

\vspace{3pt}
\noindent\textbf{Compound perturbation.}
Under compound perturbation \emph{Decomposed} reaches 80.4\% compared with 59.4\% for \emph{Total-U} and 44.6\% for \emph{Vanilla}. The 21.0~pp improvement is the largest across conditions. The decomposed controller activates observation recovery and dampening independently, whereas the combined baseline applies both interventions whenever either signal triggers.

\begin{figure}[!t]
\centering
\includegraphics[width=\columnwidth]{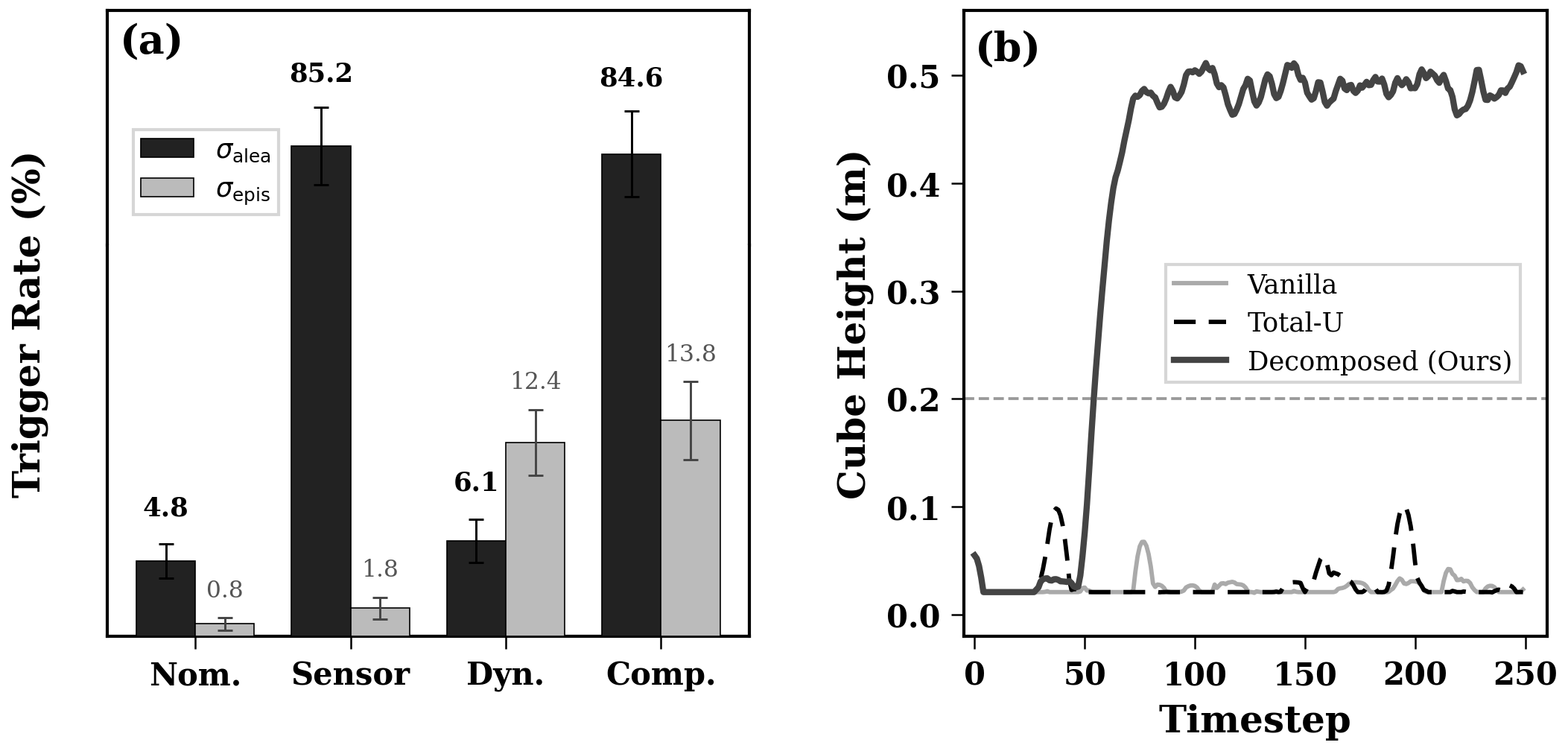}
\caption{(a)~Aleatoric and epistemic trigger rates under four perturbation conditions ($\pm1$ std over 1{,}000 episodes). $\sigma_{\text{alea}}$ fires on 85.2\% of steps under sensor perturbation while $\sigma_{\text{epis}}$ fires on 12.4\% under dynamics shift; under compound perturbation both rates persist (84.6\%, 13.8\%), confirming near-independence. (b)~Object height in a representative compound episode: \emph{Vanilla} and \emph{Total-U} fail to lift the cube above the 0.2\,m success threshold, while \emph{Decomposed} succeeds.}
\label{fig:trigger_traj}
\end{figure}

\vspace{3pt}
\noindent\textbf{Trigger rate analysis.}
Trigger frequencies explain these results. Under sensor perturbation the aleatoric signal activates on 85.2\% of timesteps while the epistemic signal activates on only 1.8\%. The controller therefore applies observation recovery without dampening. Under dynamics shift the epistemic signal activates on 12.4\% of steps while the aleatoric signal activates on 6.1\%, close to the nominal false positive rate of 4.8\%. The controller therefore applies targeted dampening while leaving observations unchanged.

\vspace{3pt}
\noindent\textbf{Asymmetry of trigger profiles.}
The contrast in Fig.~\ref{fig:trigger_traj}(a) reflects the distinct temporal structure of the two perturbation types. Sensor noise corrupts every observation with equal probability, so $\sigma_{\text{alea}}$ exceeds its threshold on 85.2\% of timesteps. Dynamics shift, by contrast, only produces prediction error when the policy interacts with modified physics --- primarily during grasp acquisition and sustained holding --- yielding a 12.4\% epistemic trigger rate that reflects when mismatch is physically observable. Under compound perturbation both rates persist (84.6\%, 13.8\%), confirming near-independence due to noise-augmented training (Eq.~\eqref{eq:ensemble_train}). \emph{Total-U} applies both interventions whenever either signal fires, producing cross-contamination that limits its compound performance to 59.4\% vs.\ 80.4\% for \emph{Decomposed} (Fig.~\ref{fig:trigger_traj}(b)).

\begin{table}[t]
\caption{Success rate improvement of \emph{Decomposed} over \emph{Total-U} in percentage points.
Results are shown for different dynamics shifts and dampening strengths.
Isaac-Lift-Cube-Franka with $N{=}1000$ episodes per condition.}
\label{tab:robo_gap}
\centering
\small
\setlength{\tabcolsep}{6pt}
\renewcommand{\arraystretch}{1.1}
\begin{tabular}{lccc}
\toprule
\textbf{Shift} & \textbf{Sensor} & \textbf{Dynamics} & \textbf{Compound} \\
\midrule
Mass $1.5\times$     & +13.6 & +10.8 & +19.5 \\
Mass $2\times$       & +13.6 & +14.1 & +21.0 \\
Friction $0.5\times$ & +13.8 & +12.0 & +20.1 \\
Friction $1.5\times$ & +14.1 & +10.0 & +19.9 \\
\midrule
\multicolumn{4}{c}{\textbf{Mass $2\times$ with varying dampening $\alpha$}} \\
\midrule
$\alpha=0.15$ & +12.0 & +11.0 & +15.9 \\
$\alpha=0.30$ & +13.6 & +14.1 & +21.0 \\
$\alpha=0.50$ & +17.8 & +15.3 & +25.0 \\
\bottomrule
\end{tabular}
\end{table}

\begin{table}[t]
\caption{Success rate (\%) under Mass $2\times$ dynamics shift for dampening strength $\alpha \in \{0.15,0.30,0.50\}$.
Isaac-Lift-Cube-Franka with $N{=}1000$ episodes per condition.
Bold indicates the higher value between methods.}
\label{tab:robo_alpha}
\centering
\small
\setlength{\tabcolsep}{6pt}
\begin{tabular}{ccccc}
\toprule
\textbf{$\alpha$} & \textbf{Method} & \textbf{Sensor} & \textbf{Dynamics} & \textbf{Compound} \\
\midrule
\multirow{2}{*}{0.15} 
& Total-U  & 82.4 & 71.8 & 63.2 \\
& Decomposed & \textbf{94.4} & \textbf{82.8} & \textbf{79.1} \\
\midrule
\multirow{2}{*}{0.30} 
& Total-U  & 80.6 & 70.1 & 59.4 \\
& Decomposed & \textbf{94.2} & \textbf{84.2} & \textbf{80.4} \\
\midrule
\multirow{2}{*}{0.50} 
& Total-U  & 76.3 & 68.2 & 54.8 \\
& Decomposed & \textbf{94.1} & \textbf{83.5} & \textbf{79.8} \\
\bottomrule
\end{tabular}
\end{table}

\subsection{Ablation Studies}

\noindent\textbf{Generalization across dynamics shifts.}
Table~\ref{tab:robo_gap} reports the success improvement of \emph{Decomposed} over \emph{Total-U} across four dynamics perturbations. Two consistent patterns appear. First, performance under sensor perturbation remains stable across shift types, with success rates near 64\% for \emph{Vanilla}, 80\% for \emph{Total-U}, and 94\% for \emph{Decomposed}. This indicates that sensor noise is handled independently of the paired dynamics condition. Second, \emph{Total-U} consistently underperforms \emph{Vanilla} under pure dynamics shift across all cases. The gap ranges from $-$2.3\% to $-$3.4\% across all four conditions, indicating that the degradation arises from the combined trigger rather than a specific physical change. Under compound perturbation the advantage of \emph{Decomposed} over \emph{Total-U} ranges from +19.5\% to +21.0\%, with larger gains under stronger dynamics shifts.

\vspace{3pt}
\noindent\textbf{Sensitivity to dampening strength.}
Table~\ref{tab:robo_alpha} evaluates dampening gains $\alpha \in \{0.15, 0.30, 0.50\}$ under the Mass $2\times$ shift. For \emph{Total-U}, increasing $\alpha$ produces monotonic degradation. Compound success decreases from 63.2\% to 59.4\% to 54.8\%, while sensor performance decreases from 82.4\% to 76.3\%. Larger dampening reduces actuation authority even when the disturbance arises from corrupted observations. Consequently no single $\alpha$ performs well across both perturbation types. In contrast, \emph{Decomposed} remains stable. Sensor success is nearly constant across the full range (94.4, 94.2, 94.1), and compound success varies by only 1.3\% (79.1 to 80.4). Independent gating isolates dampening to dynamics shifts, preventing cross-condition degradation.

\vspace{3pt}
\noindent\textbf{Runtime calibration.}
Without runtime calibration, the epistemic threshold derived from offline data activates on more than 98\% of timesteps under nominal execution. Closed-loop rollouts induce state-action distributions that differ from the independently sampled transitions in $\mathcal{D}_{\text{cal}}$, increasing prediction error by approximately $5$ to $10\times$. Runtime calibration corrects this mismatch by adjusting the threshold using a short nominal rollout. This reduces nominal trigger rates below 1\% while preserving sensitivity to dynamics shifts.

\section{Uncertainty-Guided Tracking Model Selection}
\label{sec:tracking}

Beyond corrective control, the same uncertainty signals can regulate computational allocation during perception. This section evaluates whether the proposed decomposition can guide adaptive model selection during tracking inference, avoiding unnecessary compute when uncertainty arises from measurement noise rather than model insufficiency.

\subsection{Experimental Setup}

\noindent\textbf{Datasets.}
We evaluate on two pedestrian tracking benchmarks: MOT17~\cite{milan2016mot16} and DanceTrack~\cite{sun2022dancetrack}. MOT17 contains diverse indoor and outdoor scenes with moderate crowd density, while DanceTrack features uniform appearance but complex motion patterns. Together these datasets expose both measurement noise and representation shifts.

\vspace{3pt}
\noindent\textbf{Detection models.}
Five YOLOv8 detectors with increasing capacity are available for selection: Nano (3.2M parameters), Small (11.2M), Medium (25.9M), Large (43.7M), and XLarge (68.2M). Larger models provide stronger representation capacity but incur higher computational cost.

\vspace{3pt}
\noindent\textbf{Uncertainty signals.}
Aleatoric and epistemic uncertainties are computed using the estimators from Sec.~III. Aleatoric uncertainty reflects observation noise affecting detection features, while epistemic uncertainty captures representation mismatch between the model and the visual input. These signals form the decision variables for adaptive model selection.

\vspace{3pt}
\noindent\textbf{Adaptive model selection.}
At each frame an RL controller selects one of the five detectors. The state includes the current detection confidence, $(\sigma_{\text{alea}}, \sigma_{\text{epis}})$, their temporal differences, simple spatial cues, and the active model index. The reward penalizes model capacity while assigning positive credit only when escalation reduces epistemic error or when de-escalation preserves prediction confidence. This objective encourages asymmetric behavior. Epistemic spikes trigger model scaling, whereas aleatoric uncertainty alone does not.

\subsection{Results}

\noindent\textbf{Compute–accuracy tradeoff.}
Table~\ref{tab:model_hopping} summarizes results across seven MOT17 sequences (4{,}746 frames). The adaptive policy reduces average compute by 58.2\% relative to always using YOLOv8-XLarge. The average active model size decreases from 68.2M parameters to 28.3M parameters while maintaining nearly identical detection quality. Mean IoU differs by only 0.4\% from the XLarge baseline.

Model usage concentrates on the Medium and Large tiers, which together account for 59.4\% of frames. The XLarge model activates only during short bursts of representation shift (8.6\% of frames), demonstrating that high-capacity inference is required only intermittently.

\begin{figure}[!t]
    \centering
    \includegraphics[width=\columnwidth]{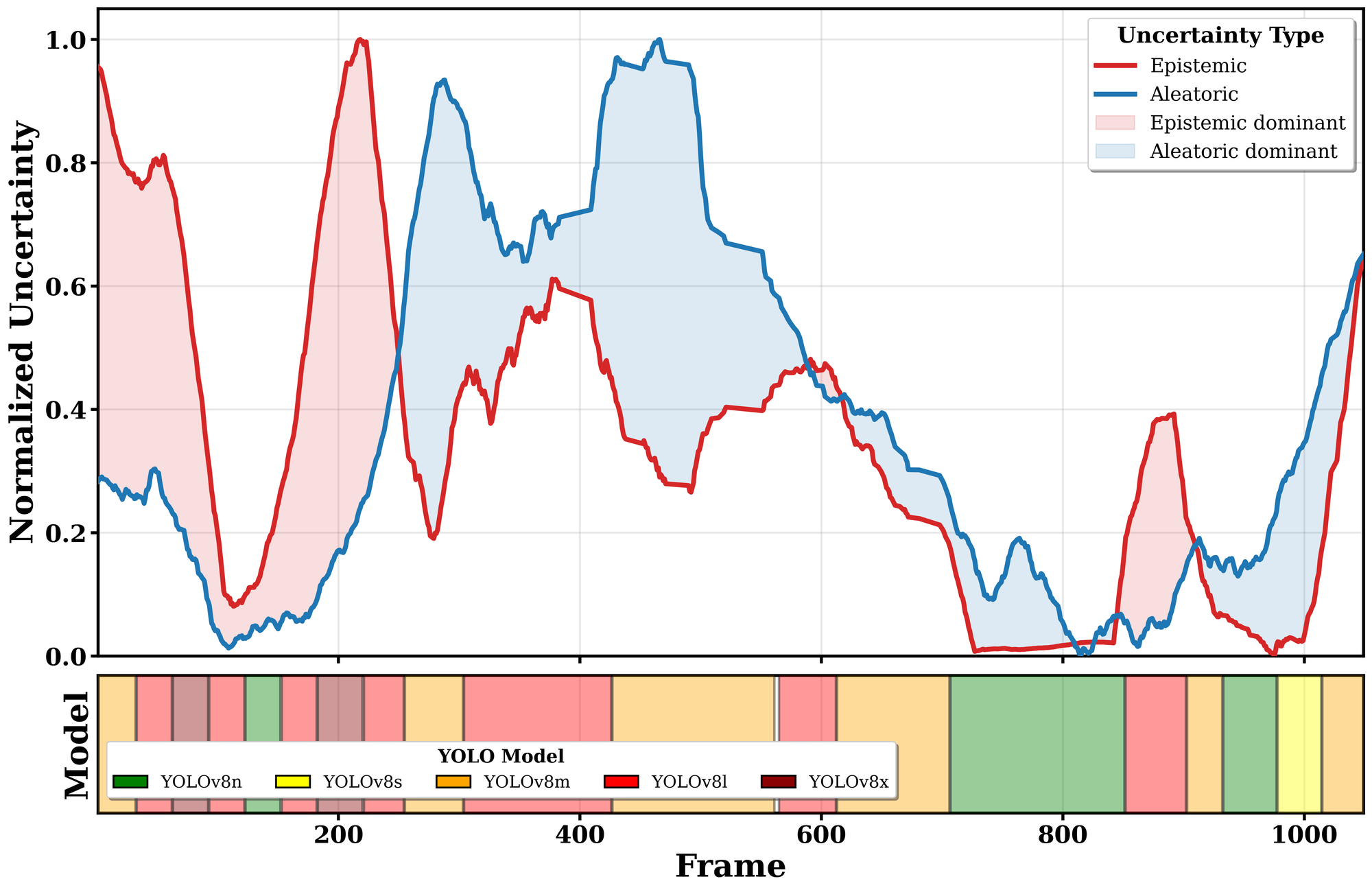}
    \caption{Adaptive model selection for MOT17-04. Top: Temporal uncertainty evolution (epistemic in red, aleatoric in blue). Bottom: Selected models color-coded by capacity (green=Nano, yellow=Small, orange=Medium, red=Large, dark red=XLarge). Policy correlates model scaling with epistemic spikes while ignoring aleatoric elevation, validating orthogonality-aware learning.}
    \label{fig:model_hopping}
\end{figure}

\begin{figure}[!t]
    \centering
    \includegraphics[width=0.75\columnwidth]{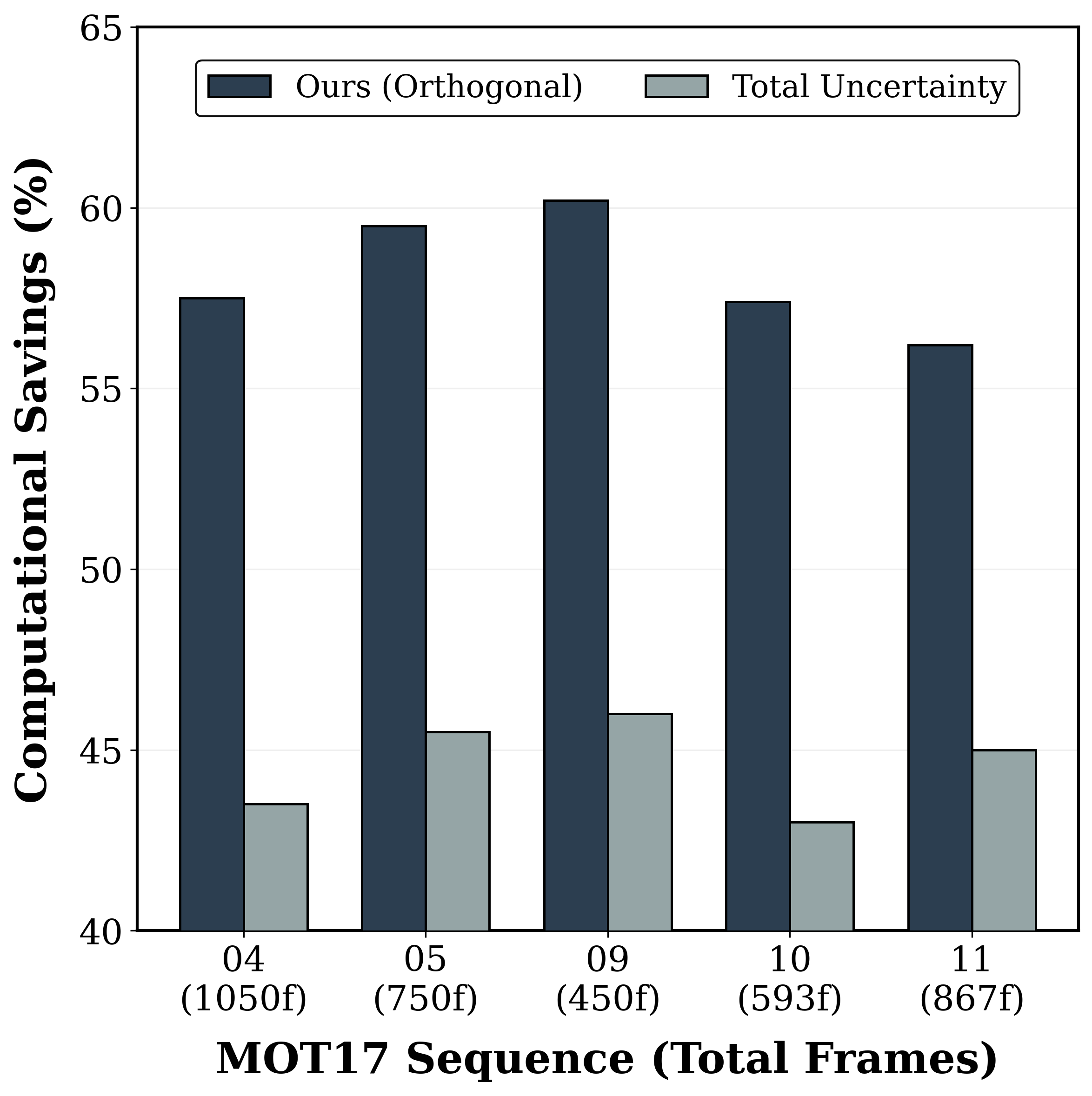}
    \caption{Ablation study comparing our orthogonal uncertainty decomposition against total uncertainty baseline across five MOT17 sequences. Our method achieves 58.2\% average computational savings vs. 44.6\% for total uncertainty (+13.6\% improvement). This substantial gap validates the necessity of separating aleatoric and epistemic components: total uncertainty conservatively uses larger models when combined uncertainty is high, while our method recognizes that high aleatoric uncertainty alone does not require increased model capacity.}
    \label{fig:ablation}
\end{figure}

\begin{table}[t]
\centering
\footnotesize
\setlength{\tabcolsep}{1pt}
\caption{Adaptive model selection across seven MOT17 sequences. The RL controller reduces compute by 58.2\% relative to always using YOLOv8-XL while preserving tracking accuracy.}
\label{tab:model_hopping}
\begin{tabular}{lcccccccc}
\toprule
\textbf{Seq} &
\textbf{Frames} &
\textbf{N (\%)} &
\textbf{S (\%)} &
\textbf{M (\%)} &
\textbf{L (\%)} &
\textbf{XL (\%)} &
\textbf{Switches} &
\textbf{Savings (\%)} \\
\midrule
02 & 550 & 18.2 & 7.3 & 35.3 & 32.0 & 7.2 & 12 & \textbf{58.3} \\
04 & 1050 & 25.2 & 4.3 & 28.1 & 34.7 & 7.7 & 18 & \textbf{57.5} \\
05 & 750 & 27.0 & 8.0 & 29.9 & 28.1 & 7.0 & 15 & \textbf{59.5} \\
09 & 450 & 26.8 & 7.1 & 29.7 & 32.4 & 4.0 & 14 & \textbf{60.2} \\
10 & 593 & 20.6 & 15.5 & 25.3 & 25.3 & 13.3 & 18 & \textbf{57.4} \\
11 & 867 & 24.7 & 6.9 & 27.8 & 27.0 & 13.6 & 19 & \textbf{56.2} \\
13 & 675 & 23.4 & 8.5 & 31.2 & 29.3 & 7.6 & 16 & \textbf{58.6} \\
\midrule
\textit{Mean} &
\textit{704} &
\textit{23.7} &
\textit{8.2} &
\textit{29.6} &
\textit{29.8} &
\textit{8.6} &
\textit{16} &
\textit{\textbf{58.2}} \\
\bottomrule
\end{tabular}
\end{table}

\vspace{3pt}
\noindent\textbf{Learned switching behavior.}
Fig.~\ref{fig:model_hopping} illustrates model selection on MOT17-04. Epistemic spikes coincide with viewpoint changes, occlusion events, and abrupt illumination shifts. During these intervals the controller escalates to larger detectors. In contrast, elevated aleatoric uncertainty caused by blur or measurement noise does not trigger escalation. Model transitions are sparse and temporally aligned with epistemic discontinuities, indicating that the controller responds to representation mismatch rather than noise fluctuations.

\vspace{3pt}
\noindent\textbf{Comparison with aggregated uncertainty.}
We compare the proposed decomposition with a baseline that uses a single combined uncertainty score. Fig.~\ref{fig:ablation} reports compute savings across five MOT17 sequences. The decomposed uncertainty controller achieves 57–60\% compute reduction, whereas the combined uncertainty baseline remains in the 43–46\% range. Aggregated uncertainty conservatively escalates model capacity whenever total uncertainty increases, even when the increase is caused by aleatoric noise. By separating the two components, the proposed method increases model capacity only when representation uncertainty rises.

\vspace{3pt}
\noindent\textbf{Compute–quality frontier.}
The adaptive policy lies above the fixed-capacity Pareto frontier, achieving 58.2\% compute reduction with only 0.4\% change in detection quality. This result confirms that decomposed uncertainty enables input-adaptive computation that responds specifically to representation mismatch rather than measurement noise.

\section{Conclusion}

This work presented a \textit{post hoc} framework that decomposes uncertainty in robotic systems into aleatoric and epistemic components and uses these signals to guide system responses. Aleatoric uncertainty reflects observation corruption and triggers observation recovery, while epistemic uncertainty reflects dynamics or representation mismatch and regulates control adaptation or model capacity. Experiments show that this separation improves robustness and efficiency across different robotic subsystems. In manipulation, the decomposed controller improves task success from 59.4\% to 80.4\% under compound perturbations and avoids the degradation observed when uncertainty is aggregated into a single scalar. In visual tracking, the same signals guide adaptive detector selection, reducing compute by 58.2\% on MOT17 while maintaining detection quality within 0.4\% of the largest backbone. These results indicate that preserving the source of uncertainty enables responses that remain aligned with the underlying disturbance, improving both corrective control and inference-time resource allocation. The principle of structured uncertainty decomposition extends naturally to other domains, including trajectory-level risk assessment in agentic reasoning systems~\cite{tayebati2026tracer}.

\FloatBarrier

{
    \small
    \bibliographystyle{IEEEtran}

}

\end{document}